\documentclass[review]{elsarticle}

\usepackage{lineno,hyperref}
\modulolinenumbers[5]
\usepackage{subfigure}
\usepackage{multirow}
\usepackage{amssymb}
\usepackage{amsmath}
\usepackage{amstext}
\usepackage{amsgen}
\usepackage{textcomp}
\usepackage{amsbsy}
\usepackage{amsopn}
\usepackage{mathrsfs}
\usepackage{tikz}
\usepackage{balance}
\usepackage{hyperref}
\usepackage{multirow}
\usepackage{colortbl}
\usepackage{xcolor}

\definecolor{Gray}{gray}{0.85}

\usepackage{todonotes}

\usepackage{multirow}
\usepackage{breqn}
\usepackage{lipsum}
\usepackage{booktabs}
\usepackage{verbatim}
\usepackage{nicefrac}
\usepackage{upgreek}
\usepackage{xfrac}
\usepackage{bbm}
\usepackage[algo2e,ruled,vlined]{algorithm2e}

\newcommand{\beq}{\begin{equation}}
\newcommand{\eeq}{\end{equation}}
\newcommand{\tr}[1]{{#1}^\top}
\renewcommand{\vec}[1]{\mathbf{#1}}
\newcommand{\mr}[1]{\mathrm{#1}}
\newcommand{\tx}[1]{\textrm{#1}}

\newcommand{\ttimes}{\,$\times$\,}

\newcommand{\loss}{\mathcal{L}}
\newcommand{\lossLab}{\loss_\mr{lab}}
\newcommand{\lossReg}{\loss_\mr{reg}}
\newcommand{\lossPen}{\loss_\mr{pen}}
\newcommand{\lossAug}{\loss_\mr{aug}}

\newcommand{\img}{x}
\newcommand{\seg}{s}
\newcommand{\labels}{y}
\newcommand{\mult}{u}
\newcommand{\seeds}{\Omega}

\newcommand{\vImg}{\vec{\img}}
\newcommand{\vSeg}{\vec{\seg}}
\newcommand{\vLabels}{\vec{\labels}}
\newcommand{\vMult}{\vec{\mult}}

\newcommand{\segReg}{\widehat{\labels}}
\newcommand{\segSize}{\widetilde{\labels}}
\newcommand{\multReg}{\widehat{\mult}}
\newcommand{\multSize}{\widetilde{\mult}}

\newcommand{\vSegReg}{\widehat{\vLabels}}
\newcommand{\vSegSize}{\widetilde{\vLabels}}
\newcommand{\vMultReg}{\widehat{\vMult}}
\newcommand{\vMultSize}{\widetilde{\vMult}}

\newcommand{\muReg}{\widehat{\mu}}
\newcommand{\muSize}{\widetilde{\mu}}

\newcommand{\params}{\uptheta}
\newcommand{\data}{\mathcal{D}}
\newcommand{\batch}{\mathcal{B}}
\newcommand{\constr}{\mathcal{C}}
\newcommand{\sizeMin}{S_\mr{min}}
\newcommand{\sizeMax}{S_\mr{max}}

\newcommand{\tskip}{\mkern1mu}

\newcommand{\visRes}[1]{\includegraphics[width=.185\textwidth]{Figs/VisRes/#1/GT.png} & \includegraphics[width=.185\textwidth]{Figs/VisRes/#1/Penalty.png} & \includegraphics[width=.185\textwidth]{Figs/VisRes/#1/CRF.png} & \includegraphics[width=.185\textwidth]{Figs/VisRes/#1/Size.png} & \includegraphics[width=.185\textwidth]{Figs/VisRes/#1/CRF_Size.png}
}

\newcommand{\visEvol}[1]{\includegraphics[width=.24\textwidth]{Figs/Evolution/GT_#1.png} & \includegraphics[width=.24\textwidth]{Figs/Evolution/CRF_Proposal_#1.png} & \includegraphics[width=.24\textwidth]{Figs/Evolution/Size_Proposal_#1.png} & \includegraphics[width=.24\textwidth]{Figs/Evolution/Network_Prediction_#1.png}
}

\journal{Neural Networks}

\begin{document}

\begin{frontmatter}

\title{Discretely-constrained deep network for weakly supervised segmentation}

\author[logti]{Jizong Peng\corref{mycorrespondingauthor}}
\author[gpa]{Hoel Kervadec}
\author[logti]{Jose Dolz}
\author[gpa]{Ismail Ben Ayed}
\author[gpa]{Marco Pedersoli}
\author[logti]{Christian Desrosiers}

\cortext[mycorrespondingauthor]{Corresponding author}
\ead{jizong.peng.1@etsmtl.ca}

\address[logti]{Department of Software and IT Engineering}
\address[gpa]{Department of Automated Production \\[1mm]
ETS Montreal, 1100 Notre-Dame W., Montreal, H3C 1K3, Canada}

\begin{abstract}
An efficient strategy for weakly-supervised segmentation is to impose constraints or regularization priors on target regions. Recent efforts have focused on incorporating such constraints in the training of convolutional neural networks (CNN), however this has so far been done within a continuous optimization framework. Yet, various segmentation constraints and regularization can be modeled and optimized more efficiently in a discrete formulation. This paper proposes a method, based on the alternating direction method of multipliers (ADMM) algorithm, to train a CNN with discrete constraints and regularization priors. This method is applied to the segmentation of medical images with weak annotations, where both size constraints and boundary length regularization are enforced. Experiments on two benchmark datasets for medical image segmentation show our method to provide significant improvements compared to existing approaches in terms of segmentation accuracy, constraint satisfaction and convergence speed.
\end{abstract}


\end{frontmatter}

% \linenumbers

\section{Introduction}

Semantic image segmentation is a fundamental problem in computer vision, which requires assigning the proper category label to each pixel of a given image \cite{liao2013representation,milletari2016v}. This problem is essential to various applications such as autonomous driving \cite{luc2017predicting} and neuroimaging \cite{yuan2017automatic,dolz20173d}. Among the wide range of segmentation systems, approaches based on deep convolutional neural network (CNN) have recently attracted great attention \cite{dolz20173d,szegedy2017inception,litjens2017survey}. While providing state-of-art performance in many segmentation tasks, CNN-based approaches usually require a large training set of fully-annotated images that can be both time-consuming and expensive to obtain. This limitation is especially important in medical imaging where access to patient data is restricted and annotation requires expert-level knowledge.

Higher-level (or weak) annotations, such as image-level tags \cite{pinheiro2015weakly,papandreou2015weakly,kervadec2018constrained,pathak2015constrained}, bounding boxes \cite{dai2015boxsup,rajchl2017deepcut} and scribbles \cite{kervadec2018constrained,lin2016scribblesup}, can generally be obtained more efficiently than pixel-wise annotations. This type of weak supervision has therefore been intensively investigated for segmentation. Another valuable strategy in a weakly-supervised setting is to use prior knowledge to guide the segmentation of unlabeled or partly-labeled images during training \cite{kervadec2018constrained,pathak2015constrained}. This is particularly useful for medical segmentation problems, where information about the target region is often known beforehand. %MP: in my understanding when using weakly supervised or semi-sup. you always need some prior infos, thus to me weak sup. and prior knowledge are the same thing... 

While these works have reduced the performance gap between CNNs trained with fully-annotated images and those trained in a weakly-supervised setting, they have explored constrained optimization from a continuous perspective. %MP: maybe if you have space, explain more the meaning of continuous perspective, that is looking at probabilities
Yet, many constraints, such as bounds on the size of segmented regions, are better expressed discretely. In continuous methods, the size of a region is typically estimated by computing the sum of pixel probabilities corresponding to this region. However, limiting this sum does not guarantee that the actual size of the region, after thresholding probabilities, will meet constraints. Similarly, regularization priors for segmentation like boundary length \cite{boykov2001fast}, star-shapeness \cite{veksler2008star} and compactness \cite{dolz2017unbiased} are usually discrete, and optimizing them in a continuous framework is susceptible to local minima. Lastly, solving sub-problems in a discrete manner, instead of using gradient descent, benefits from globally-optimal algorithms which can significantly improve the current solution in a single update step.

The contribution of this work is two-fold: 
\begin{itemize}
\item We present a first method to train a CNN with discrete constraints and regularization priors. The proposed method uses an efficient strategy, based on the alternating direction method of multipliers (ADMM) algorithm, to separate the optimization of network parameters with SGD from optimizing discretely-constrained segmentation labels. We show that updating these discrete labels can be done in polynomial time with guarantee of solution optimality. 
\item We apply the proposed method for the segmentation of medical images with weak annotations. While previous works have considered either size constraints \cite{kervadec2018constrained,pathak2015constrained} or boundary length regularization \cite{bai2017semi} for segmentation, our method combines these two priors in a single efficient model. Experiments on three segmentation problems show our method to yield significant improvements compared to existing approaches in terms of segmentation accuracy, constraint satisfaction and convergence speed.
\end{itemize}

The rest of this paper is organized as follows. In Section~\ref{sec:rel} we give an overview of related work. We then present our discretely-constrained model in Section~\ref{sec:methodology} and describe the experiments to evaluate its performance in Section~\ref{sec:exp}. Finally, we give experimental results in Section~\ref{sec:results} and draw conclusions in Section~\ref{sec:conclusion}.

\section{Related work}
\label{sec:rel}

\subsection{Constrained segmentation} 

Several works have tackled the problem of weakly-supervised segmentation by imposing constraints on deep CNNs \cite{kervadec2018constrained,pathak2015constrained,kervadec2019curriculum,jia2017constrained,zhou2019prior}. In \cite{pathak2015constrained}, Pathak et al. propose a latent distribution and KL-divergence to constrain the output of a segmentation network. Their method allows decoupling the optimization of network parameters with stochastic gradient descent (SGD) from the update of the latent distribution under constraints. It is used in a semi-supervised setting to impose size constraints and image-level tags (i.e., force the presence or absence of given labels) on the regions of unlabeled images. Moreover, a simple $L_2$ penalty term was proposed in \cite{jia2017constrained} to impose equality constraints on the size of the target regions in the context of histopathology image segmentation. More recently, Kervadec et al. \cite{kervadec2018constrained} showed that imposing inequality constraints on size directly in gradient-based optimization, also via an $L_2$ penalty term, provided better accuracy and stability than the approach in \cite{pathak2015constrained} when few pixels of an image are labeled. %Even though both methods employ a similar $L_2$ penalty term, the equality-constrained formulation in \cite{jia2017constrained} requires exact knowledge of the target size, while the inequality-constrained formulation in \cite{kervadec2018constrained} allows much more flexibility as to the required prior domain-specific knowledge. 
Similarly, Zhou et al. \cite{zhou2019prior} embedded prior knowledge on the target size in the loss function by matching the probabilities of the empirical and predicted output distributions via the KL divergence. As directly minimizing this term by standard SGD is difficult, they proposed to optimize it by using stochastic primal-dual gradient. While these works have helped improve segmentation in a weakly-supervised setting, they have mainly focused on continuous optimization methods.

%\jose{These are more weakly supervised papers:} \cite{bai2017semi,pinheiro2015weakly,papandreou2015weakly,dai2015boxsup,rajchl2017deepcut,lin2016scribblesup}

\subsection{Discrete-continuous optimization} 

Discrete optimization has a long history in research, playing a key role in various problems of computer science and applied mathematics \cite{wolsey2014integer}. Theoretical results in this field, like the minimization of submodular functions \cite{cunningham1985submodular}, have given us efficient tools to solve complex decision problem involving discrete variables. In recent years, significant efforts have been made to combine these powerful tools with optimization techniques for continuous problems. In \cite{miksik2014distributed}, an ADMM algorithm is used to perform distributed inference in large-scale Markov Random Fields (MRF), using both discrete and continuous variables in the optimization. A similar idea is proposed in \cite{dolz2017dope} to minimize discrete energy functions (submodular and non-submodular) for distributed image regularization. Other lines of work include discrete-continuous methods based on ADMM for incorporating high-order segmentation priors on the target region's histogram of intensities \cite{karnyaczki2015sparse} or compactness \cite{dolz2017unbiased}. More recently, similar techniques have been proposed to include higher-order priors directly in the learning process \cite{laude2018discrete,marin2019beyond}. To our knowledge, our work is the first employing a discrete-continuous framework for weakly-supervised segmentation.

\section{Methodology}\label{sec:methodology}

\begin{figure*}[t!]
 \centering
 %\shortstack{
 %\includegraphics[width=0.9\linewidth]{Figs/diagram.png} 
 %\\ \vphantom{1mm}}
 \includegraphics[width=.95\linewidth]{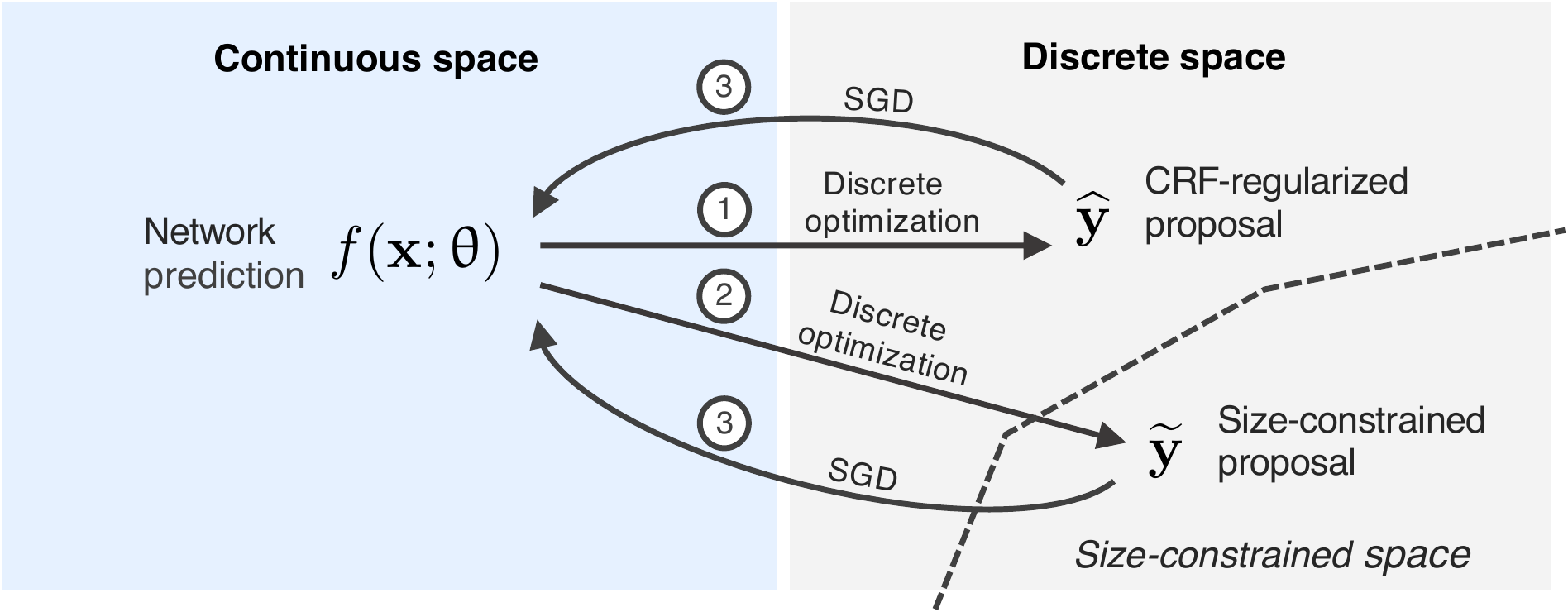} 
 \caption{Diagram of the proposed discrete-continuous model for weakly-supervised segmentation.  %\textcolor{red}{Jose: I think we may need some numbers/identifiers/etc in the input of the discriminators (blue box) to make more clear (specially to lazy reviewers) from where the inputs come from..Something like A,B,C,D}
 }
 \label{fig:diagram}
\end{figure*}

\subsection{Problem formulation}

We focus on the following weakly-supervised segmentation problem where the labels of training images are only provided for a small subset of pixels. Given a training dataset $\data = \{(\vImg^i, \vLabels^i, \Omega^i)\}_{i=1}^N$ where $\vImg^i$ is an image, $\Omega^i$ is the subset of labeled pixels, and $\vLabels^i$ the corresponding labels, we want to learn a segmentation model $f$ parameterized by $\params$, such that $f(\vImg^i,\params)$ gives the label probabilities at each pixel. For simplicity, we suppose a two-class segmentation problem and, using $\vSeg^i(\params) = f(\vImg^i; \params)$ as short-hand notation, denote as $\seg_p^i \in [0,1]$ the foreground probability predicted for a pixel $p$. 

To learn parameters $\params$ in this setting, we impose three requirements on the network output: 1) it should respect labeled pixels in $\Omega^i$; 2) it should satisfy \emph{discrete} segmentation constraints $\constr_j$, for $j=1,\ldots,M$; 3) it should minimize a \emph{discrete} regularization term $\lossReg$. Considering these requirements, we formulate the task as the following optimization problem:
\begin{equation}\label{eq:continuous_formulation}
\begin{aligned}
\min_{\params} \ & \ \loss(\params) \, = \, \sum_{i=1}^N \lossLab\left(\vSeg^i(\params), \vLabels^i\right) \ + \ \lambda\tskip\lossReg\left(\vSeg^i(\params)\right) \\
  \tx{s.t.} \ &  \ \constr_j\left(\vSeg^i(\params)\right) \, \leq \, 0,  \ \ i=1,\ldots,N, \ \ j=1,\ldots,M, 
\end{aligned}
\end{equation}
In this formulation, $\lossLab$ imposes the segmentation network to agree with provided pixel annotations, $\lossReg$ enforces the segmentation to be spatially regular, and $\lambda$ is a hyper-parameter controlling the trade-off between label satisfaction and regularization. 

The formulation in Eq. (\ref{eq:continuous_formulation}) is very general and covers a wide range of segmentation applications. For the purpose of this paper, however, we focus on a particular scenario similar to the one considered in \cite{kervadec2018constrained}. As in this previous work, we define $\lossLab$ as the cross-entropy loss over labeled pixels:
\begin{equation}
\lossLab\left(\vSeg^i, \vLabels^i\right) \, = \,  - \sum_{p \tskip \in \tskip \seeds^i} \labels_p^i \log \seg_p^i  \, + \, \big(1-\labels_p^i\big) \log \left(1-\seg_p^i\right).
\end{equation}
Moreover, while any regularization prior can be used for $\lossReg$, we consider in this work the weighted Potts model \cite{boykov2001fast}. Let $\tau(x) = \mathbbm{1}(x \geq 0.5)$ be the function converting probability $x$ to a binary value, this prior is defined as $\lossReg(\vSeg^i) = \sum_{p,q} w^i_{p,q} \, |\tau(\seg_p^i) - \tau(\seg_q^i)|$, where $w^i_{p,q} = \exp\left(-\frac{1}{2\sigma^2} |\img_p^i - \img_q^i|^2\right)$ if pixels $p$ and $q$ are within a given neighborhood of each other, else $w^i_{p,q}=0$. For neighbor pixels, the weight measures the similarity of pixel intensity or color. This regularization prior, which is frequently used for segmenting medical images, estimates the boundary length of the foreground region, hence minimizing it gives a segmentation with smooth contours. %Since anatomical regions often satisfy this property, such prior is frequently used for segmenting medical images. 
Last, as in \cite{kervadec2018constrained} and \cite{pathak2015constrained}, we impose lower and upper size bounds on the foreground: $\sizeMin \leq \sum_{p} \tau(s^i_p) \leq \sizeMax$. This type of constraint is also well-suited for medical image segmentation since the size of anatomical regions is typically limited by biology and size bounds can often be found in anatomical studies.

As we will show in our experiments, the combination of boundary regularization and size constraints on the foreground is essential to achieve good performance in the weakly-supervised setting of this work. Since we only have partially-labeled images, size constraints are generally insufficient to uniquely define the foreground. Boundary regularization helps the network focus on regions that also respect edges in the image. However, since the contrast between the foreground and background varies largely from one image to another, using only  this regularization with a fixed parameter $\lambda$ can lead to under- or over-segmentation (e.g., the network predicting foreground or background at every pixel). Hence, adding size constraints to the regularization prior provides a way to control the effect of this prior on individual images. 

In the next sections, we present two popular approaches, the penalty method and the ADMM algorithm, for solving constrained optimization problems like the one in Eq. (\ref{eq:continuous_formulation}). We then explain how the latter can be used to solve this problem efficiently with a mixed continuous-discrete formulation.

\subsection{Penalty-based optimization}\label{sec:penalty}

A popular method for constraining the output of a neural network is to model the constraint as a penalty term in the loss function \cite{kervadec2018constrained,pathak2015constrained}. This additional loss term is typically a differentiable, convex function which equals zero for any output satisfying the constraint, and otherwise produces a positive value proportional to the degree of constraint violation. The main advantage of this approach is that it can be used directly within standard optimization techniques for training neural networks, for instance stochastic gradient descent (SGD), and typically leads to a smooth optimization (e.g., little oscillation due to the constraint). However, the penalty-based approach also suffers from important limitations which we illustrate by considering the simple problem of weakly-supervised segmentation with only an upper bound on the foreground size, i.e. $\sum_p s^i_p(\params) \leq \sizeMax$. Using a squared penalty function, the problem can be expressed as
\begin{equation}\label{eq:penalty_formulation}
\begin{aligned}
\lossPen(\params) \, = \, \sum_{i=1}^N \lossLab\left(\vSeg^i(\params), \vLabels^i\right) \ + \ \frac{\mu}{2}\Big(\max\Big\{0, \, \sum_p \seg^i_p(\params) - \sizeMax\Big\}\Big)^2.
\end{aligned}
\end{equation}
For the task at hand, the above formulation suffers from three important problems. First, it requires tuning penalty parameter $\mu$, possibly for each image, otherwise the constraint may not be satisfied. Second, since the foreground size is estimated as the sum of probability values, instead of hard label assignments, it is very likely that the constraint will not be met once probability values are thresholded. For example, assigning a 50\% foreground probability to all pixels in the image gives the same sum as giving a 100\% probability to half the pixels. A possible strategy to enforce hard assignments in the network is to increase the temperature parameter of the softmax, however this leads to gradient saturation which freezes the solution in a local minima. The third problem can be understood by looking at the gradient of the loss with respect to network parameters, i.e.
\begin{equation}\label{eq:penalty_grad}
\begin{aligned}
\nabla_{\params}\lossPen(\params) \, = \, \sum_{i=1}^N \nabla_{\params}\lossLab\left(\vSeg^i(\params), \vLabels^i\right) \ + \  F^i \, \sum_p \nabla_{\params}\seg^i_p(\params),
\end{aligned}
\end{equation}
with $F^i = \mu \cdot \max\Big\{0, \, \sum_p \seg^i_p(\params) - \sizeMax\Big\}$. We see that, if the foreground size is greater than $\sizeMax$, the network simply scales down the gradient for each pixel by a constant factor $F^i$ (note that the actual change in probability for a pixel is also proportional to its prediction uncertainty). This uniform scaling of gradient can result in a bad local minima if the shape to segment is complex (e.g., curved and narrow like the right ventricle) or the initial network output is poor.

\subsection{Optimization with ADMM}

Because the regularization prior and size constraints are discrete, the formulation in Eq. (\ref{eq:continuous_formulation}) cannot be optimized directly. To alleviate this problem, we propose and approach based on the alternating direction method of multipliers (ADMM) algorithm \cite{boyd2011distributed}. ADMM is a variant of the augmented Lagrangian scheme which uses partial updates for the dual variables. It is often employed to solve problems in the form of $\min_{\vImg} f(\vImg) + g(\vImg)$, where optimizing functions $f$ and $g$ together is hard, however minimizing each of them separately can be done more easily (e.g., to optimality and/or efficiently). The main idea is to reformulate the task as a constrained optimization problem $\min_{\vImg,\vLabels} f(\vImg) + g(\vLabels)$ subject to $\vImg\!=\!\vLabels$, for which the objective function is separable in $\vImg$ and $\vLabels$, and solve this new problem using an augmented Lagrangian method
\begin{equation}
    \max_{\vMult} \, \min_{\vImg, \vLabels} \ \ \lossAug(\vImg,\vLabels,\vMult) \ = \ f(\vImg) \, + \, g(\vLabels) \, + \, \tr{\vMult}(\vImg-\vLabels) \, + \, \frac{\mu}{2}\left\|\vImg-\vLabels\right\|^2,
\end{equation}
where $\vMult$ are the Lagrange multipliers and $\mu$ is the ADMM penalty parameter. An alternate definition, which we will employ in this paper, uses scaled multipliers \cite{boyd2011distributed}:
\begin{equation}\label{eq:scaled_admm}
    \max_{\vMult} \, \min_{\vImg, \vLabels} \ \ \lossAug(\vImg,\vLabels,\vMult) \ = \ f(\vImg) \, + \, g(\vLabels) \, + \, \frac{\mu}{2}\left\|\vImg - \vLabels + \vMult\right\|^2.
\end{equation}
Optimization of Eq. (\ref{eq:scaled_admm}) is performed by solving iteratively with respect to each variable, while keeping others fixed, until convergence is reached:
\begin{align}
    \vImg^{t+1} & \ := \ \mr{argmin}_{\vImg} \    
        \lossAug(\vImg,\vLabels^{t},\vMult^{t}) \nonumber\\
    \vLabels^{t+1} & \ := \ \mr{argmin}_{\vLabels} \
        \lossAug(\vImg^{t+1},\vLabels,\vMult^{t}) \\
    \vMult^{t+1} & \ := \ 
        \vMult^{t} \, + \, (\vImg^{t+1}-\vLabels^{t+1}) \nonumber
\end{align}
The ADMM algorithm can be used to efficiently solve certain types of constrained optimization problems. Thus, we can suppose that function $g$ models a hard constraint, i.e. it returns $0$ if its input satisfies the constraint, else it returns infinity. The main advantage of this formulation, compared to the penalty-based method in Eq. (\ref{eq:penalty_formulation}), is simplicity: we can minimize $f$ in a unconstrained problem (update of $\vImg$), and finding a feasible solution amounts to solving a simple proximal problem (update of $\vLabels$). As shown in the next section, this simplicity enables the use of efficient techniques in the discrete optimization setting. Another benefit of ADMM is that it dynamically adjusts the strength of the penalty, using Lagrange multipliers, to enforce the satisfaction of constraints. Hence, it is less sensitive to the choice of $\mu$ than the penalty method.

\subsection{Discretely-constrained segmentation using ADMM}\label{sec:our_method}

We now show how to use ADMM to solve the problem of Eq. (\ref{eq:continuous_formulation}) efficiently. We propose a mixed discrete-continuous formulation, where network parameters are optimized using standard gradient-based optimization, while the size-constrained segmentation problem is solved using specialized techniques for discrete optimization. Toward this goal, we decouple the label loss $\lossLab$, discrete regularization loss $\lossReg$ and discrete size constraint by introducing two binary segmentation vectors $\vSegReg^i$ and $\vSegSize^i$ for each training image $\vImg^i$, and adding the following equality constraints: $\vSeg^i(\params) = \vSegReg^i$, $\vSeg^i(\params) = \vSegSize^i$. These added vectors can be thought of as segmentation proposals, which are updated iteratively from the network's predictions, and help transfer discrete regularization and constraints to the training process. As we later present, this particular variable splitting strategy is chosen so that updating these proposals can be done efficiently and to optimality. Figure \ref{fig:diagram} summarizes the proposed strategy.

Using the ADMM formulation of Eq. (\ref{eq:scaled_admm}), we then rewrite the segmentation problem as
\begin{equation}\label{eq:admm}
\begin{aligned}
\max_{\vMultReg, \tskip \vMultSize} \ \min_{\params, \tskip \vSegReg, \tskip \vSegSize}  \ & \ \lossAug\left(\params, \vSegReg, \vSegSize, \vMultReg, \vMultSize\right) \, = \,  
    \sum_{i=1}^N \lossLab\left(\vSeg^i(\params), \vLabels^i\right) \ + \ \lambda\tskip\lossReg(\vSegReg^i) \\[-1.5mm]
       & \qquad \ + \ \frac{\muReg}{2}  \left\| \vSeg^i(\params) - \vSegReg^i + \vMultReg^i\right\|_2^2 
  \ + \, \frac{\muSize}{2}  \left\| \vSeg^i(\params) - \vSegSize^i  + \vMultSize^i\right\|_2^2 \\[1mm]
\tx{s.t.} & \ \ \sizeMin \, \leq \, \sum_{p} \segSize_p^i \, \leq \, \sizeMax,  \ \forall i\\[-2mm]
& \ \ \vSegReg^i, \, \vSegSize^i \, \in \, \{0,1\}^{n_i}, \ \forall i.
\end{aligned}
\end{equation}
Here, $n_i$ is the number of pixels in image $\vImg^i$. In the following subsections, we explain how each variable of this problem is updated (Lagrange multipliers are updated as per the standard ADMM method).

\subsubsection{Network parameters} 
\label{sec:network_params}

To update network parameters, we use a mini-batch gradient descent technique. Let $\batch \subset \data$ be a batch of training samples, the gradient of the loss for batch $\batch$ is given by 
\begin{equation}\label{eq:params_gradient}
%\frac{\partial}{\partial\params}  \ : \  
    \sum_{i \tskip \in \tskip \batch} \nabla_{\params} \lossLab(\vSeg^i, \vLabels^i) \ + \ \sum_p \left( \muReg(\seg^i_p - \segReg^i_p + \multReg^i_p) 
      \, + \, \muSize(\seg^i_p - \segSize^i_p + \multSize^i_p) \right) \nabla_{\params}\seg^i_p(\params).
\end{equation}  
We see that, unlike the penalty method gradient in Eq. (\ref{eq:penalty_grad}), the gradient at each pixel of image is scaled in a non-uniform manner based on the discrete CRF-regularized and size-constrained proposals. The parameters are then updated by taking a step opposite to the gradient, i.e. $\params^{t+1} \, := \, \params^{t} - \eta \nabla_{\params} \lossAug^t$, where $\nabla_{\params}\lossAug$ is the gradient defined in Eq. (\ref{eq:params_gradient}) and $\eta$ is the learning rate. One should note that this gradient-based update does not guarantee an optimal solution for the network parameters, which is normally required for ADMM. Instead, this corresponds to a proximal variant of ADMM in which a first-order approximation is used for the objective function \cite{he2017convergence,zhang2018proximal}.

\subsubsection{CRF-regularized proposal}
\label{sec:crf_proposal}

Considering all other variables fixed, updating each CRF-regularized proposals $\vSegReg^i$ amounts to solving 
\begin{equation}
 \min_{\vSegReg^i \tskip \in \tskip \{0,1\}^{n_i}} \  \frac{1}{2}  \left\| \vSegReg^i -(\vSeg^i(\params) + \vMultReg^i)\right\|_2^2 \, \ + \ (\sfrac{\lambda}{\muReg})\,\lossReg(\vSegReg^i).   
\end{equation}
Using the property that $x^2 = x$ for a binary variable $x$, this problem can be expressed equivalently as a standard CRF energy minimization problem
\begin{equation}\label{eq:crf-reg}
 \min_{\vSegReg^i \tskip \in \tskip \{0,1\}^{n_i}} \  \sum_{p} \widehat{a}^i_p \, \segReg_p^i \ + \ (\sfrac{\lambda}{\muReg})\,\lossReg(\vSegReg^i),   
\end{equation}
where $\widehat{a}^i_p = \tfrac{1}{2} - \seg_p^i(\params) - \multReg_p^i$ are unary potentials and $\lossReg$ is a pairwise (or higher-order) regularization prior. If $\lossReg$ is a sub-modular function, as the weighted Potts model used in this work, then the global optimum of this discrete optimization problem can be obtained in polynomial time with a max-flow algorithm \cite{boykov2001fast}. Note that this would not be the case if foreground size constraints were added to Eq. (\ref{eq:crf-reg}), which motivates the splitting strategy chosen for our method. 

\subsubsection{Size-constrained proposal}
\label{sec:size_proposal}

Likewise, we update each proposal $\vSegSize^i$  by considering all other variables fixed and solving the following constrained discrete problem: 
\begin{equation}
 \min_{\vSegSize^i \tskip \in \tskip \{0,1\}^{n_i}} \  \frac{1}{2}  \left\| \vSegSize^i -(\vSeg^i(\params) + \vMultSize^i)\right\|_2^2, \quad
 \tx{s.t.} \ \ \sizeMin \, \leq \, \sum_{p} \segSize_p^i \, \leq \, \sizeMax.   
\end{equation}
Using the same trick for binary variables as before, we then rewrite this problem as
\begin{equation}\label{eq:size_problem} 
    \max_{\vSegSize^i \tskip \in \tskip \{0,1\}^{n_i}}  \ \sum_{p} \widetilde{a}^i_p\,\segSize_p^i, \quad
 \tx{s.t.} \ \ \sizeMin \, \leq \, \sum_{p} \segSize_p^i \, \leq \, \sizeMax,
\end{equation} 
with $\widetilde{a}^i_p = \seg_p^i(\params) - \multSize_p^i - \tfrac{1}{2}$. This discrete problem corresponds to a specific instance of the knapsack problem \cite{chu1998genetic}, where each pixel $p$ is an object with equal weight $1$ and utility $\widetilde{a}^i_p$. The goal is to select between $\sizeMin$ and $\sizeMax$ objects such that their total utility is maximized. The optimal solution to this problem can be obtained via a simple ranking method where we set $\segSize_p^i=1$ for the $\sizeMin$ pixels with highest utility and, if any, for the remaining pixels with highest \emph{positive} utility until $\sizeMax$ is reached. Table \ref{tab:examples_ranking} illustrates this ranking procedure on a toy example.

\begin{table}[t!]
\begin{small}
\begin{center}
\begin{tabular}{rccccccc}
%$p$ & 1 & 2 & 3 & 4 & 5 & 6 \\
%\hline
Ranked $\widehat{a}_p$ & : & 0.95 & 0.88 & 0.52 & -0.11 & -0.64 & -0.79 \\
\hline
$\sizeMin=2$, $\sizeMax=5$ & : & 1 & 1 & 1 & 0 & 0 & 0 \\
$\sizeMin=3$, $\sizeMax=5$ & : & 1 & 1 & 1 & 0 & 0 & 0 \\
$\sizeMin=4$, $\sizeMax=5$ & : & 1 & 1 & 1 & 1 & 0 & 0
\end{tabular}
\end{center}
\end{small}
\vspace*{-2mm}
\caption{Illustration of the ranking method for update the size-constrained proposal $\vSegSize$. The top row gives pixel utility values, ranked in decreasing order. Following rows gives the corresponding proposal for different lower ($\sizeMin$) and upper ($\sizeMax$) size bounds.} \label{tab:examples_ranking}
\end{table}

%\begin{spacing}{0.8}

\begin{center}
\SetAlFnt{\small\sffamily}
\begin{algorithm2e}[t!]
%\setstretch{1.35}
\SetNoFillComment
%
%\nocaptionofalgo
\KwIn{Weakly-labeled images $\data = \{(\vImg^i, \vLabels^i, \Omega^i)\}_{i=1}^N$;}    
\KwIn{Size bounds $[\sizeMin, \sizeMax]$;}
\KwOut{Network parameters $\params$;}
\caption{Discretely-constrained segmentation}\label{algo}

\BlankLine
\tcc{Initialization}
Randomly initialize network parameters $\params$\;
Set $\segReg^i_p, \, \segSize^i_p = \sfrac{1}{2}$ and $\multReg^i_p, \, \multSize^i_p = 0$, $\forall i, p$\;

\BlankLine
\tcc{Main loop}
\For{$\mr{epoch} = 1, \ldots, E_{\mr{max}}$}{
    \BlankLine
    \tcc{Network parameters update}             
    \For{$\mr{iter} = 1, \ldots, T_{\mr{max}}$}{
        Randomly select batch $\batch \subset \data$\;
        Apply batch gradient step as in Section \ref{sec:network_params}\;
    }    
    \BlankLine
    \tcc{Discrete proposals update}
    Update CRF-regularized proposal $\vSegReg^i$ as in Section \ref{sec:crf_proposal}\;
    \mbox{Update size-constrained proposal $\vSegSize^i$ as in Section \ref{sec:size_proposal}}\;
    \BlankLine    
    \tcc{Multipliers update}
    $\vMultReg^i := \vMultReg^i \, + \, (\vSeg^i(\params) - \vSegReg^i)$, $\forall i$\;
    $\vMultSize^i := \vMultSize^i \, + \, (\vSeg^i(\params) - \vSegSize^i)$, $\forall i$\;
    \BlankLine    
    Decrease learning rate: $\eta := \mr{decr}_{\eta}\times\eta$, with $\mr{decr}_{\eta} \in [0,1]$\;
} 

\BlankLine
\Return{$\params$} \;
\end{algorithm2e}
\end{center}
%\end{spacing}

\subsection{Algorithm summary and complexity}

The whole training process is summarized in Algorithm \ref{algo}. Starting with zero-valued multipliers and proposals with equal foreground and background probabilities for each pixel, each training epoch involves the following steps. First, for $T_{\mr{max}}$ iterations, we update network parameters by randomly selecting a batch of 2D training images, computing the gradient for this batch and applying a descent step with this gradient. Next, we re-compute the 3D CRF-regularized proposals ($\vSegReg^i$) and size-constrained proposals ($\vSegSize^i$) using the modified network output. Last, we update Lagrange multipliers for both proposals and reduce the learning rate by a factor of $\mr{decr}_{\eta}$. We note that the network doesn't need to be re-trained from scratch after each proposal update and that the only requirement for convergence is that the update of network parameters decreases the overall loss \cite{boyd2011distributed}.

Since proposals are updated only once per epoch, our method yields negligible computational overhead compared to optimizing only the  network with SGD. Moreover, each of these updates has low computational complexity. Thus, computing each $\vSegReg^i$ is done by solving a max-flow problem, which has $\mathcal{O}(n^3)$ complexity where $n$ is the number of image pixels/voxels. Likewise, updating each $\vSegSize^i$ simply requires to sort pixels/voxels, the complexity of which is in $\mathcal{O}(n \log n)$. 

\section{Experiments}
\label{sec:exp}

\subsection{Datasets and evaluation protocol}
\label{datasets}

We evaluate the proposed method on three different medical imaging segmentation tasks: left-ventricular (LV) and right-ventricular (RV) endocardium segmentation in cine magnetic resonance imaging (MRI), and prostate segmentation in T2-MRI.\\[2mm]
\noindent\textbf{LV and RV segmentation:} This medical image set is provided by the Automated Cardiac Diagnosis Challenge (ACDC) \cite{bernard2018deep} and focuses on the segmentation of three cardiac structures, i.e. left ventricular endocardium and epicardium, and right ventricular endocardium. It consists of 100 cine MRI exams covering normal cases and subjects with well-defined defined pathologies: dilated cardiomyopathy, hypertrophic cardiomyopathy, myocardial infarction with altered left ventricular ejection fraction and abnormal right ventricle. Each exam contains acquisitions at the diastolic and systolic phases. The spatial resolution goes from 0.83 to 1.75 mm$^2$/pixel, with a thickness of 5-8 mm and an inter-slice gap of 5 mm, covering the LV from the base to the apex. In our experiments, 80 exams were employed for training and the remaining 20 for validation.

%\chris{From Hoel's paper} The exams were acquired in breath-hold with a retrospective or prospective gating and a SSFP sequence in 2-chambers, 4-chambers and in short-axis orientations. A series of short-axis slices cover the LV from the base to the apex, with a thickness of 5-8 mm and an inter-slice gap of 5 mm. The spatial resolution goes from 0.83 to 1.75 mm$^2$/pixel. For our experiments, we employed 75 exams for training and the remaining 25 for validation.\\[2mm]
\noindent\textbf{Prostate segmentation:} 
For the third task, we used the dataset made available at the MICCAI 2012 prostate MR segmentation challenge (PROMISE) \cite{litjens2014evaluation}. This dataset contains multi-centric transversal T2-weighted MR images from 50 subjects acquired with multiple MRI vendors and different scanning protocols, which are representative of typical MR images acquired in a clinical setting. The images resolution ranges from 15\ttimes256\ttimes256 to 54\ttimes512\ttimes512 voxels with a spacing ranging from 2\ttimes0.27\ttimes0.27 to 4\ttimes0.75\ttimes0.75 mm$^3$. In this case, we employed 40 patients for training and 10 for validation during the experiments.

\begin{figure}[t!]
  \begin{center}
  \begin{small}
  \setlength{\tabcolsep}{1pt}
  \begin{tabular}{cccc}
  %\rowcolor{Gray}
    \includegraphics[width=0.24\textwidth]{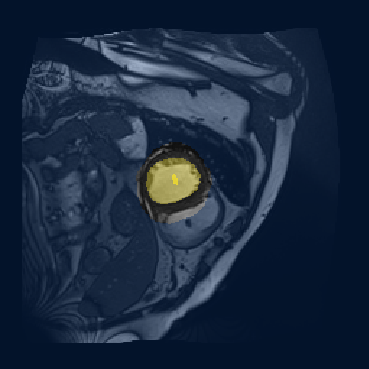}  &    
    \includegraphics[width=0.24\textwidth]{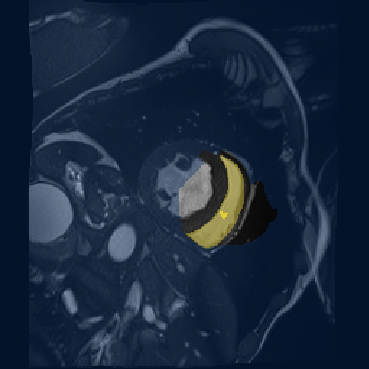} &    
    \includegraphics[width=0.24\textwidth]{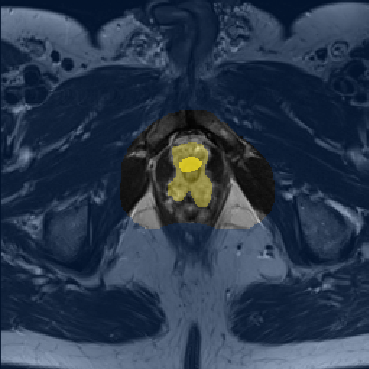}\\[-1mm]
    %\rowcolor{Gray}
    \includegraphics[width=0.24\textwidth]{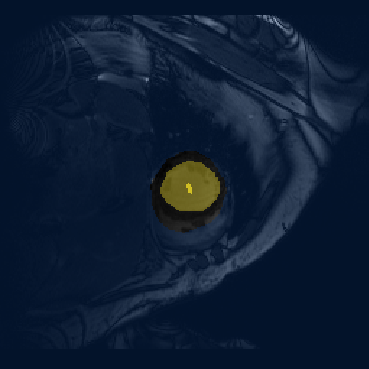}  &    
    \includegraphics[width=0.24\textwidth]{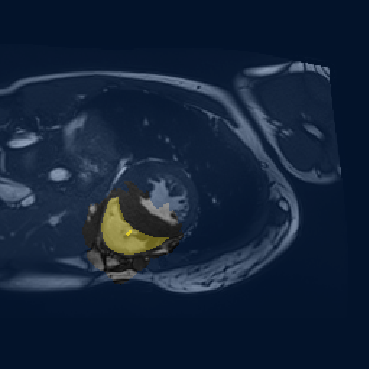} &    
    \includegraphics[width=0.24\textwidth]{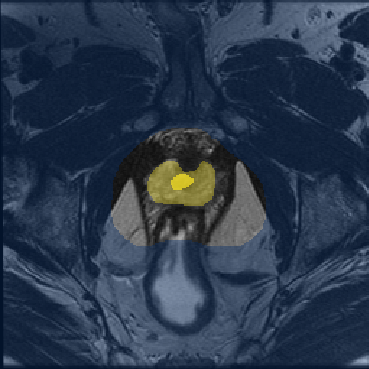}\\%[-1mm]
    LV (train) & RV (train) & Prostate (train)
    \end{tabular}   
    \end{small}
  \caption{Examples of images, ground truth and annotations for the three tested datasets. Bright yellow pixels correspond to foreground seeds, light yellow pixels to the ground-truth, and blue-shaded pixels to the background mask. Note that we only show a single slice of a 3D volume, and many slices (further away from the foreground center) actually contain no foreground seed.}\label{fig:examples_datasets}
  \end{center}
\end{figure}

We generate weak annotations based on the scenario where an annotator identifies for each volume a single position (e.g., mouse click) near the centroid of the foreground region, and a global anatomical atlas is then used to obtain foreground and background seeds from this position. The following procedure is employed to derive annotations under this scenario. Considering training images only, we compute the centroid of each 3D ground truth mask and use it as the origin of a global coordinate system, i.e. position $(0,0,0)$ in each volume corresponds to the foreground centroid in that volume. For each position in this new coordinate system, we then measure the fraction of training volumes for which the position is in the foreground region. Next, we define a global foreground mask containing all positions having a foreground probability of 100\%, and a global background mask containing all those with 0\% probability.
%Both foreground and background masks are then shrunk by applying a 3D erosion operator with kernel size of 2 voxels. 
Finally, annotations are obtained for each training volume by aligning the global masks to the volume's foreground centroid, and using the aligned masks to label voxels in the volume accordingly. %Note that different probability tresholds could be employed to define the foregound and background masks (e.g., $\geq$95\% and $\leq$5\%), however this would result in incorrect annotations.

Figure \ref{fig:acdc_train} shows examples of annotations and ground truth for the three test datasets. It can be seen that the proposed procedure for generating weak annotations leads to diverse segmentation tasks with various levels of difficulty. For left ventricle segmentation (LV), foreground seeds represent on average 14.29\% of the foreground region and 0.12\% of the whole volume. In comparison, they only represent 2.29\% of the foreground and 0.02\% of the volume for right ventricle segmentation (RV). Moreover, for non-circular regions like RV, we see that foreground seeds can be close to the region boundary, making the segmentation more challenging. 

We test three different settings for the proposed method, using 1) only CRF-regularized proposals, 2) only size-constrained proposals, and 3) both CRF-regularized and size-constrained proposals. For the two settings using size-constrained proposals, we define size bounds of different tightness by adding or substracting a relative percentage $\epsilon \in \{0\%, 10\%, 20\%, 40\%\}$ of the real foreground size: $[\sizeMin,\sizeMax] = [(1-\epsilon)\!\times\!S_{\mr{real}}, \, (1+\epsilon)\!\times\!S_{\mr{real}}]$. Hence, $\epsilon=0$ imposes the predicted segmentation to have the same size as the ground truth. 

To get an upper bound on performance, we trained our segmentation model using fully-annotated images and call this baseline \emph{fully-supervised} in the results. We also test the penalty approach \cite{kervadec2018constrained} presented in Section \ref{sec:penalty}, using the same network architecture, seeds and partial cross-entropy loss as for our method. As described before, this approach estimates the foreground size by summing the probabilities predicted for this region over the whole image, and then applies a squared-error loss on the difference between the estimated size and the lower or upper size bounds. 

The performance of segmentation models is evaluated using the 3D Dice similarity coefficient (DSC). This well-known metric measures the overlap between the predicted segmentation $S$ and ground truth mask $G$, as
\begin{equation}
    \mr{DSC}(S,G) = \ \frac{2|S \cap G|}{\,\,|S| + |G|}.
\end{equation} 

\subsection{Implementation details}

For all experiments on the ACDC dataset, we followed \cite{kervadec2018constrained} and used ENet \cite{paszke2016enet} as our segmentation architecture. This lightweight network, which has been used in urban segmentation and medical imaging, gives a good trade-off between accuracy and inference speed. We trained the network from scratch with SGD using the Adam optimizer and a batch size of 8. As in \cite{kervadec2018constrained}, 3D volumes are segmented slice-by-slice using images of size 256\ttimes256 as input to the network, without data augmentation. The initial learning rate was set to 5\ttimes10$^{-4}$ and decreased by a factor of 4 every 50 epochs for a total of 250 training epochs. %An cross-entropy loss is used for our proposed method to impose FG and BG seeds on the prediction. We measured and reported the DSC performance on the validation set.
For CRF regularization, we used $\lambda=100$ for all experiments and selected a different $\sigma$ to account for contrast differences between the left-ventricle (LV) and right-ventricle (RV): $\sigma=10^{-4}$ for LV and $\sigma=10^{-5}$ for RV.

For the PROMISE dataset, we employed the Unet architecture with 15 layers, batch normalization and ReLU activation. This architecture is one of the most popular models for segmentation, especially for tasks related to bio-medical imaging. While batch normalization was not used in the original work, our experiments showed it to significantly improve training speed and stability. We trained the network from scratch using the Adam optimizer with a weight decay of $10^{-4}$ and a batch size of 4. The initial learning rate was set to $10^{-4}$ and decreased by a factor of 5 every 50 epoch for a total of 250 epochs. Since this segmentation task is more challenging, to have a good performance for the fully-supervised baseline, we performed the following data augmentation procedure: images resized to 256\ttimes256 pixels, followed by random rotation, random crop, and random flip. CRF regularization parameters were set to $\lambda=1000$ and $\sigma=10^{-6}$.

The same value was employed for ADMM penalty parameters of our CRF\,+\,size method (i.e., $\muReg = \muSize$). This value, as well as the value of the penalty method parameter $\mu$, were selected for each segmentation task using grid search. Unless specified otherwise, we report for all methods the results obtained using the best value found for these parameters.

%We evaluate the segmentation performance on the validation set by using weakly-annotated labels as well as size constraints and/or smoothness priors. By providing an adaptive size estimation for each image, we achieve to boost the performance significantly. In addition, the smoothness prior imposed for the output of the CNN prediction can further improve the segmentation performance, reaching similar performance of fully-supervised training with pixel-wisely annotated labels. Experimental results validate the effectiveness of discrete optimization to help CNN training and show that our proposed method can effectively close the gap between models employing fully-annotated dataset and weakly-annotated ones. 

\section{Results}\label{sec:results}

In this section, we evaluate the performance of the proposed strategy for weakly-supervised segmentation and compare it to the penalty method in \cite{kervadec2018constrained}, which is the closest to our work. We also assess the benefit of using CRF-regularized and size-constrained proposals for training the network, and measure segmentation accuracy for different size bounds. Last, we evaluate the impact of the ADMM penalty parameter on results. 

\begin{table}[t!]
\begin{center}
 \begin{small}
\begin{tabular}{llccc}
\toprule
& \bf{Method} & \bf{LV} & \bf{RV} & \bf{Prostate} \\
\midrule\midrule
& Fully-supervised & 0.927 & 0.870 & 0.873\\
\midrule
& Ours (CRF only) & 0.862 & 0.677 & 0.769\\
\midrule
\multirow{3}{*}{\shortstack{Size bound:\\[1.5mm] $(\epsilon=0\%)$}} 
 & Penalty \cite{kervadec2018constrained} & 0.813 & 0.529 & 0.588 \\
 & Ours (size only) & 0.871 & 0.598 & 0.760 \\
 & Ours (CRF\,+\,size) & \bf{0.901} & \bf{0.730} & \bf{0.807} \\
\midrule
\multirow{3}{*}{\shortstack{Size bound:\\[1.5mm] $(\epsilon=10\%)$}} 
 %& Penalty \cite{kervadec2018constrained} & 0.832 & 0.486 & 0.621 \\
 & Penalty \cite{kervadec2018constrained} & 0.840 & 0.570 & 0.621 \\
 & Ours (size only) & 0.844 & 0.617  & 0.766 \\
 & Ours (CRF\,+\,size) & \bf{0.884} & \bf{0.719} & \bf{0.795} \\
\midrule
\multirow{3}{*}{\shortstack{Size bound:\\[1.5mm] $(\epsilon=20\%)$}} 
 & Penalty \cite{kervadec2018constrained} & 0.833 & 0.498& 0.583\\
 %& Penalty \cite{kervadec2018constrained} & 0.824 & 0.573& --\\
 & Ours (size only) & 0.845 & 0.600 & 0.760\\
 & Ours (CRF\,+\,size) & \bf{0.872} & \bf{0.734} & \bf{0.809}\\ 
\midrule
\multirow{3}{*}{\shortstack{Size bound:\\[1.5mm] $(\epsilon=40\%)$}} 
 & Penalty \cite{kervadec2018constrained} & 0.826 & 0.582 & 0.515 \\ 
 %& Penalty \cite{kervadec2018constrained} & 0.824 & 0.582 & -- \\
 & Ours (size only) & 0.827 & 0.570  & 0.758\\
 & Ours (CRF\,+\,size) & \bf{0.879} & \bf{0.691} & \bf{0.771} \\ 
\bottomrule
\end{tabular}
\end{small}
\end{center}
\caption{Mean 3D Dice of tested methods on validation images for left ventricle (LV) and right ventricle (RV) segmentation of the ACDC dataset and prostate segmentation of the PROMISE dataset. For the penalty method and our method using size proposals (size only or with CRF), we report accuracy for different foreground size bounds defined by $\epsilon$.}
\label{tab:main_results_new}
\end{table}

\subsection{Segmentation performance}\label{sec:performance}

Table \ref{tab:main_results_new} reports the results of our discretely-constrained method under three settings, i.e. using only CRF-regularized proposals, only size-constrained proposals, or both types of proposals, comparing them to the penalty method in \cite{kervadec2018constrained} and the fully-supervised baseline. Overall, the proposed method achieves a higher mean 3D Dice than the penalty approach for the same size bounds. %, since the constraints are imposed discretely and not over the output softmax probabilities. 
This improvement ranges from 4-5\% in the case of LV to nearly 20\% for the RV and prostate segmentation tasks, and is consistent across nearly all tasks and configurations. We then examine the impact of using image-specific size constraints with different $\epsilon$, when no CRF regularization is used (i.e., size only). As bounds on the size become less restrictive (i.e., larger $\epsilon$), the segmentation performance of the proposed method degrades, resulting in a decrease of 3-4\% with respect to the setting with $\epsilon=0\%$. This indicates that an accurate size estimation can help improve segmentation in a weakly-supervised setting. 

\begin{figure}[t!]
\begin{center}
\setlength{\tabcolsep}{0pt}
\begin{small} 
  \begin{tabular}{cc}
    \includegraphics[width=0.5\textwidth]{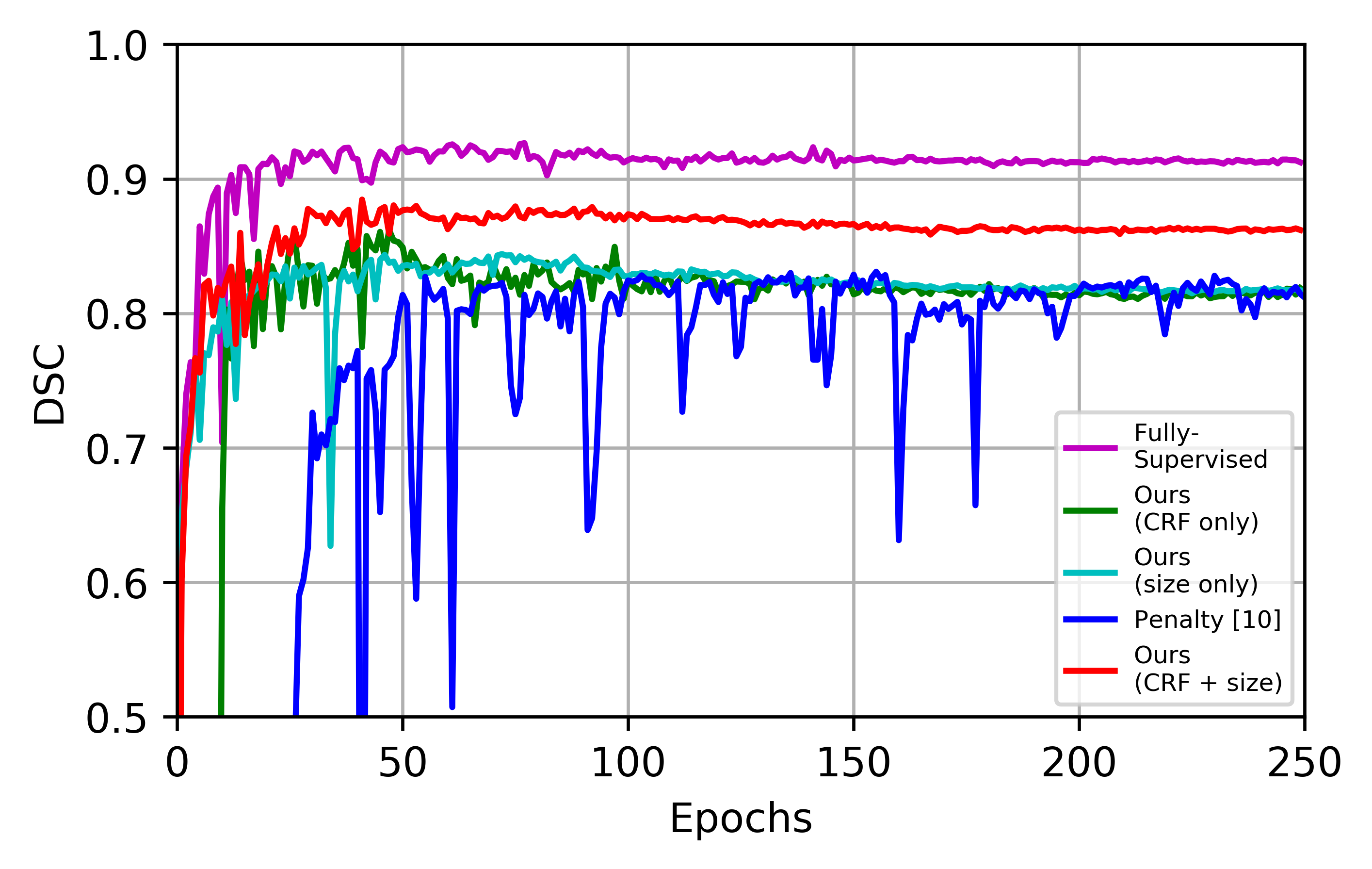} &
    \includegraphics[width=0.5\textwidth]{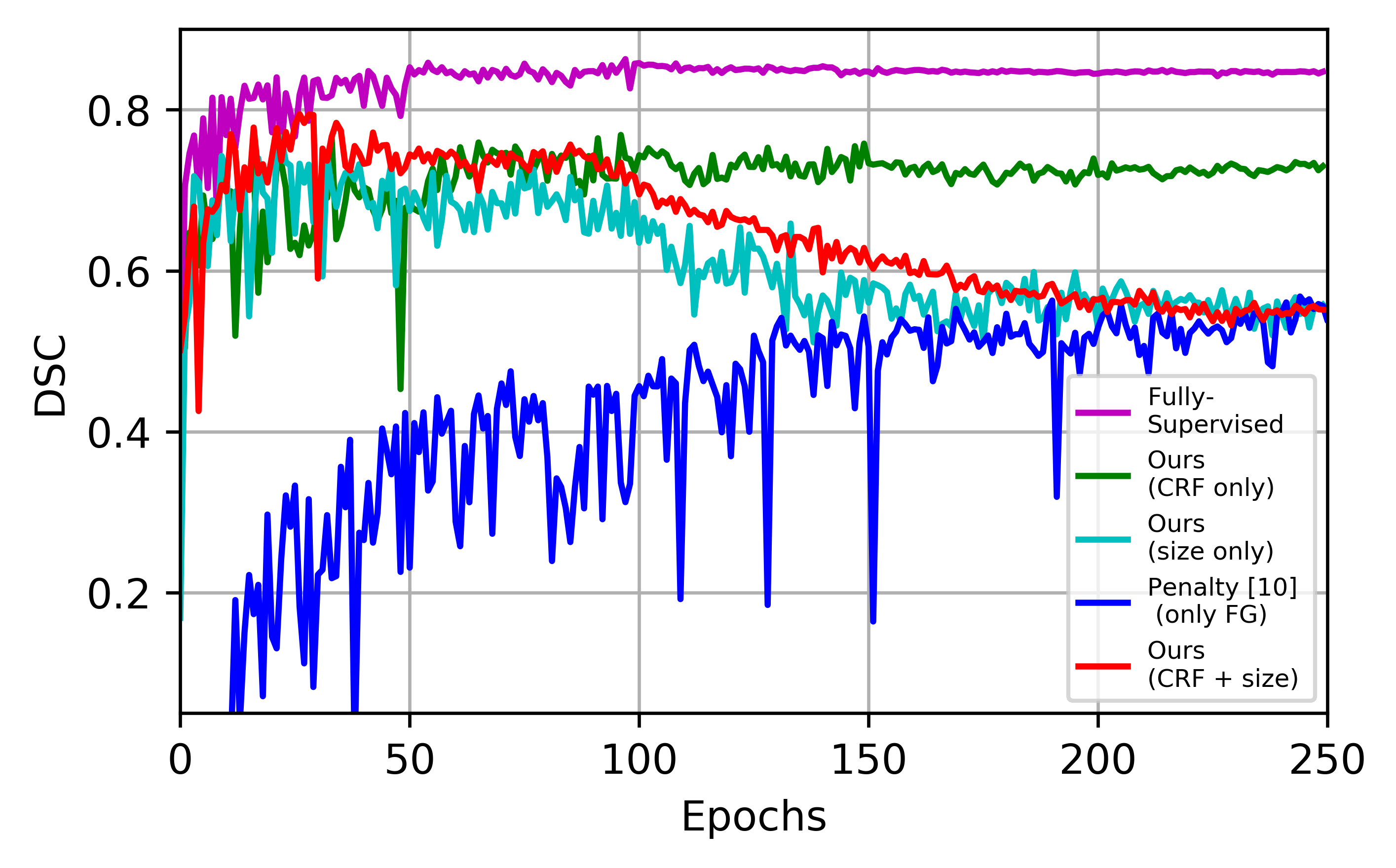} \\[-2mm]
    LV (validation) & Prostate (validation)
  \end{tabular}
\end{small}
  \caption{3D Dice on validation images of the ACDC (LV) and PROMISE prostate datasets during training. Note: foreground size bounds corresponding to $\epsilon=10\%$ are employed for the penalty method and our method using size proposals (size only or with CRF).}
  \label{fig:acdc_train}  
\end{center}  
\end{figure}

We also observe that employing a regularization prior jointly with size constraints (i.e.,  CRF\,+\,size) leads to a better performance, with DSC improvements of 3\% to 15\% compared to using only size proposals. Improvements are particularly important for the RV segmentation task which presents a more complex shape (i.e., curved and narrow) than LV or prostate, suggesting that imposing size constraints alone is not sufficient to get a satisfactory segmentation on certain structures. By including the CRF regularization, the segmentation is attracted towards the target contours, resulting in higher 3D Dice. The benefits of combining both priors can also be appreciated when comparing against the model with only CRF regularization. A drawback of CRF regularization is its sensitivity to parameter $\lambda$, which should normally be tuned per image to avoid under- or over-segmentation. Adding a size prior adds robustness to this parameter, even when size constraints are less restrictive. Particularly, results show an accuracy boost of 3-5\% over CRF-regularization alone when employing tight bounds ($\epsilon=0\%$), and a gain of 1-2\% when bounds are loosest ($\epsilon=40\%$). %This demonstrates that imposing such constraint on the target size can help to by-pass the well-known shrinking bias problem on CRFs. %Additionally, even when the size estimation is coarse --$\epsilon$ of 40\%--, the DSC can still be improved by our method to outperform the CRF baseline.  

The 3D validation Dice of tested methods during training is depicted in Fig. \ref{fig:acdc_train} for the LV and prostate segmentation tasks. Compared to the penalty strategy, the proposed method converges faster and generally shows a more stable behaviour. This underlines the importance of proposals to guide the network in the beginning of training, when images are partly labeled. Interestingly, the highest validation accuracy of our method is obtained early in the training (epoch 50).

\begin{figure}[ht!]
\begin{center}
\renewcommand{\arraystretch}{.85}
\setlength{\tabcolsep}{1pt}
\begin{small} 
  \begin{tabular}{cccc}
  \visEvol{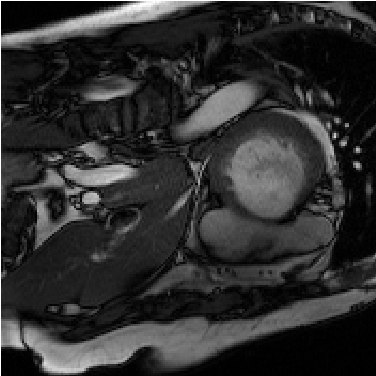} \\
  \visEvol{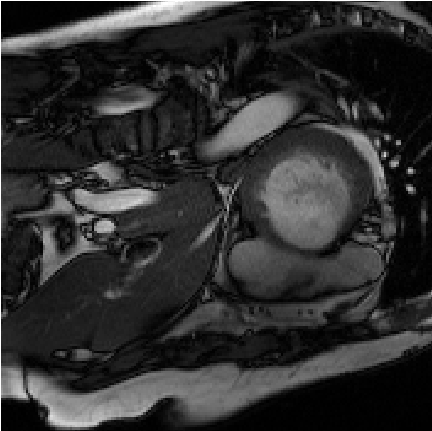} \\
  \visEvol{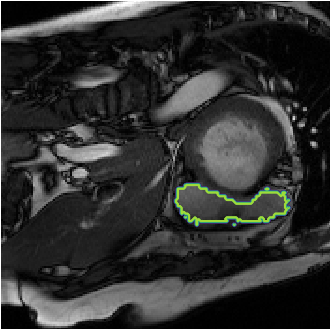} \\
  \visEvol{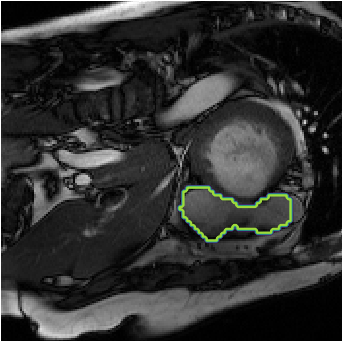} \\[-2mm]
    GT & CRF proposal & Size proposal & Network output
  \end{tabular}
\end{small}
  \caption{Evolution of the CRF-regularized proposal ($\vSegReg$), size-constrained proposal ($\vSegSize$) and network output at different training stages of our  CRF\,+\,size method for $\epsilon=0$. From top to bottom: epoch$=0$, epoch$=2$, epoch$=10$ and epoch$=250$.}
  \label{fig:evol_proposals}
\end{center}  
\end{figure}

Figure \ref{fig:evol_proposals} illustrates the evolution of the CRF-regularized proposal, size-constrained proposal and network prediction at different epochs for a training image. We observe that the CRF proposal, although not perfect, plays an important role in the beginning. As training progresses, the size proposal then helps correct errors of the CRF by constraining the foreground size. At convergence, the network prediction and proposals are nearly identical. As mentioned above, the best segmentation is obtained before convergence occurs (see epoch 10 in the figure). 

\begin{figure}[t!]
\begin{center}
\begin{small}
\renewcommand{\arraystretch}{.85}
\setlength{\tabcolsep}{1pt}
  \begin{tabular}{ccccc}
  \visRes{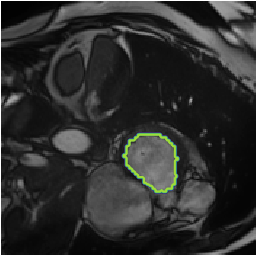} \\
  \visRes{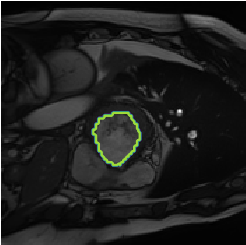} \\
  %\visRes{LV3} \\
  \visRes{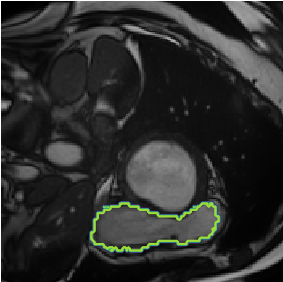} \\
  \visRes{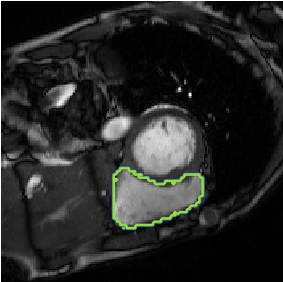} \\
  %\visRes{RV3} \\
  %\visRes{Prostate1} \\
  \visRes{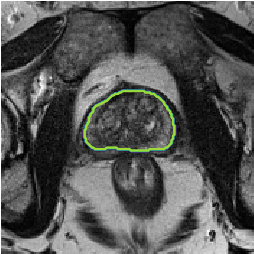} \\
  \visRes{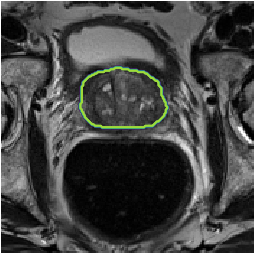} \\
  GT & Penalty \cite{kervadec2018constrained} & \shortstack{Ours \\ (CRF only)} & \shortstack{Ours \\ (size only)} & \shortstack{Ours \\ (CRF\,+\,size)}
  \end{tabular}
\end{small}  
  \caption{Visual comparison of tested methods on validation images for size bounds corresponding to $\epsilon=10\%$. Top two rows: ACDC left ventricle (LV).  Middle two rows: ACDC right ventricle (RV). Bottom two rows: PROMISE prostate.}  
  \label{fig:visualization}
\end{center}  
\end{figure}

\subsection{Qualitative evaluation}

Visual results of tested methods for the three segmentation tasks are depicted in Fig. \ref{fig:visualization}. %Visual inspection shows that the proposed strategy brings several benefits over the penalty method in \cite{kervadec2018constrained} and the standard CRF-regularization. 
We observe that incorporating only size constraints during training might be insufficient to segment complex structures in a weakly-supervised scenario, regardless of the method used. For example, in the RV segmentation task (middle two rows), both the penalty approach and our method with only size proposals produce segmentation masks which are not well aligned with the image target boundaries, although their size are similar to the ground truth. Nevertheless, imposing discrete size constraints helps generate contours whose sizes are closer to the real target size, as shown in the last row of the figure. Inspecting the contours obtained by our model using only CRF regularization, we observe that they better match the target boundaries. However, the well-known shrinking bias problem of CRFs make this model under-segment regions in some cases (e.g., see first two rows). The proposed model can overcome this issue by integrating both size and CRF regularization priors. %By doing so, it can be observed that the CRF pushes the segmentation toward the boundaries, while the size constraint prevents the model to undersegment, avoiding shrinking bias and achieving segmentation results close to the ground truth.

%Visual inspection shows that incorporating only size prior during training might be insufficient to segment complex structures in a weakly supervised scenario. For example, in the RV segmentation task, both 

 %First, we observe that even though both the proposed strategy and the penalty term approach typically produce predictions not aligned with the object boundaries, imposing size constraints discretely generates contours whose sizes are closer to the real target size. The use of CRF regularization can overcome this limitation and push the prediction towards the image boundaries. 

%...differences only size (continuous vs discrete)...improvement when CRF is included to size...compared to standard CRF

\begin{figure}[t!]
\begin{center}
\setlength{\tabcolsep}{2pt}
\begin{small} 
  \begin{tabular}{cc}
    \includegraphics[width=0.49\textwidth]{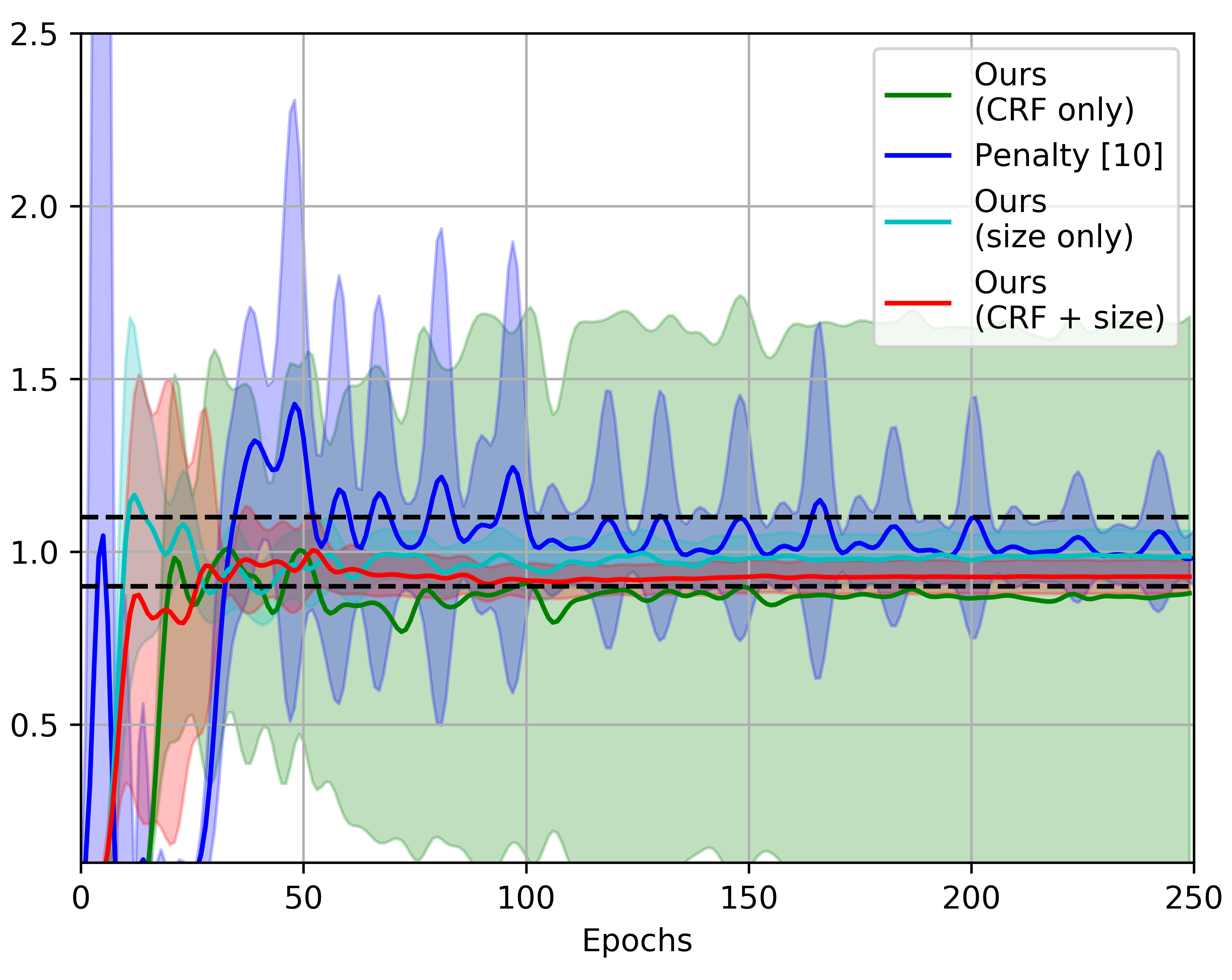} &
    \includegraphics[width=0.49\textwidth]{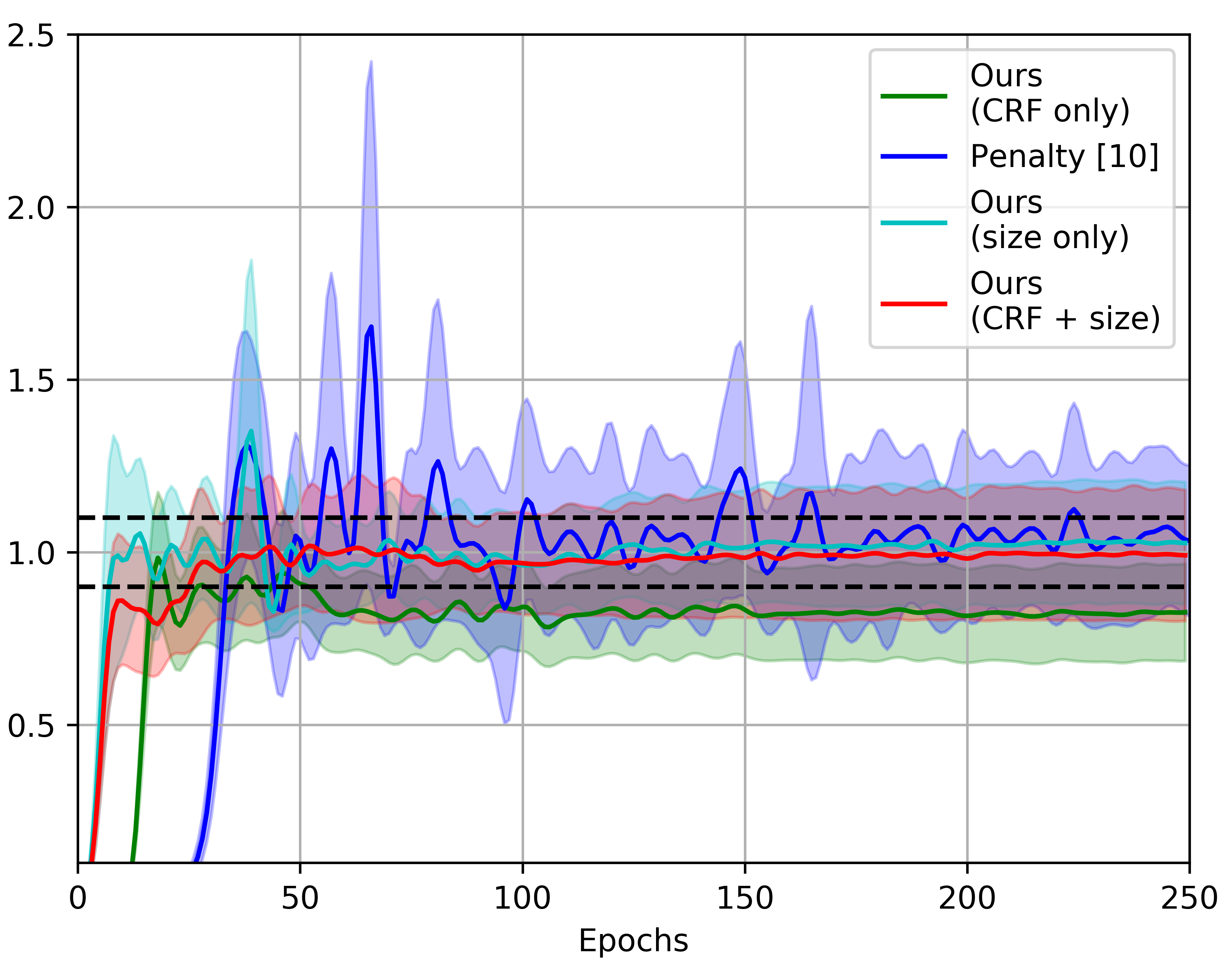} \\[-2mm]
    LV (train) & LV (validation)
  \end{tabular}
\end{small}
  \caption{Ratio between predicted and real foreground size during training, while using size bounds corresponding to $\epsilon=10\%$ (dashed lines). Solid lines correspond mean value and light-colored intervals to standard deviation.
  %\textcolor{red}{Can we change the legend to have a single line for each strategy? And do you have these plots for RV and prostate too?}
  }  
  \label{fig:size_epoch}
\end{center}  
\end{figure}

%Lastly, compared to the upper --fully supervised-- baseline, the proposed method achieves 

\subsection{Constraint satisfaction}

In Fig. \ref{fig:size_epoch}, we evaluate our method's ability to satisfy imposed constraints. We consider size bounds corresponding to $\epsilon=10\%$ and compute the ratio between the size of the predicted segmentation and the real foreground size for the different model settings. We first observe that the model with only CRF regularization (green curve) does not control the size of the foreground, as its predictions are pushed towards the visible boundaries in the image regardless of the target size. While the penalty approach (blue curve) converges to a mean ratio within size bounds, its behaviour remains unstable and generates segmentations whose size lies outside the imposed bounds, both during training and validation. On the other hand, if the size constraints are imposed in the discrete domain, the ratio between predicted and real sizes is also located within the bounds, but follows a more stable regime. This pattern is also observed in the configuration with both size and CRF priors (red curve).  However, this configuration converges faster to predictions satisfying constraints, particularly in the validation set. 

\begin{table}[t!]
\begin{center}
 \begin{small}
\begin{tabular}{lcccc}
\toprule
\bf{Method} & $\boldsymbol{\muReg \, = \, \muSize}$ & \bf{~LV~} & \bf{~RV~} & \bf{Prostate} \\
\midrule\midrule
\multirow{4}{*}{CRF only} 
 & 0.01 & 0.798 & 0.541 & 0.703 \\
 & 0.1 & 0.825 & 0.594 & 0.744 \\
 & 1 & \bf{0.862} & \bf{0.677} & \bf{0.769} \\
 & 10 & 0.828 & 0.675 & 0.768 \\
 \midrule
\multirow{4}{*}{\shortstack{Size only\\[1.5mm] ($\epsilon=0$)}} 
 & 0.01 & 0.797 & 0.548 & 0.695 \\
 & 0.1 & 0.814 & \bf{0.598} & 0.741 \\
 & 1 & \bf{0.871} & 0.576 & \bf{0.760} \\
 & 10 & 0.724 & 0.341 & 0.645 \\
 \midrule
\multirow{4}{*}{\shortstack{CRF\,+\,size \\[1.5mm] ($\epsilon=0$)}} 
 & 0.01 & 0.783 & 0.549 & 0.744 \\
 & 0.1 & 0.835 & 0.604 & 0.740 \\
 & 1 & 0.878 & 0.722 & 0.756 \\
 & 10 & \bf{0.901} & \bf{0.730} & \bf{0.807} \\
\bottomrule
\end{tabular}
\end{small}
\end{center}
\caption{Mean 3D Dice obtained with different values for the ADMM penalty parameters $\muReg$ and $\muSize$. For our CRF\,+\,size method, we used the same value for both parameter (i.e., $\muReg=\muSize$). Tight bounds on foreground were employed for these results (i.e., $\epsilon=0$).}
\label{tab:parameter_impact}
\end{table}

\subsection{Impact of the ADMM penalty parameter}

As last experiment, we measure the sensitivity of our method to the ADMM penalty parameters $\muReg$ and $\muSize$. As described in Section \ref{sec:our_method}, these parameters control the trade-off between the constraint satisfaction loss (i.e., quadratic penalty term) and the supervised loss (i.e., partial cross-entropy). For our  CRF\,+\,size method, the same value was used for both penalty parameters. Table \ref{tab:parameter_impact} gives the mean 3D Dice obtained by the three settings of our method for different values of the penalty parameter. While the optimal value varies from one setting to the other, i.e. the best value for CRF only and size only is 1 while it is 10 for CRF\,+\,size, the best value for a given setting seems stable across different segmentation tasks. 

\section{Discussion and conclusion}\label{sec:conclusion}

We presented a novel method for training a CNN with discrete constraints and regularization priors. This method uses ADMM to split the continuous optimization of network parameters with SGD from the computation of discrete segmentation proposals. By incorporating both constraints and regularization priors, the network can be trained efficiently with weak annotations. We applied the proposed method to the segmentation of cardiac structures and prostates from MRI data, using partially-labeled images and bounds on the foreground size. Experiments show our method to provide significant improvements compared to the penalty approach of \cite{kervadec2018constrained}, in terms of segmentation accuracy, constraint satisfaction and convergence speed.  

One of main advantages of our method compared to the optimization of constraints directly within a gradient-based framework is its ability to perform large steps in the solution space, with guaranteed optimality. However, this can also lead to instability during training (e.g., pixels in the discrete proposals oscillating between the foreground and background classes) if the penalty term is too strong. As described in \cite{boyd2011distributed}, a possible way to alleviate this problem is to update the ADMM penalty parameter dynamically during training, for example starting with a smaller value and increasing it over epochs. Another important advantage of decoupling the supervised loss from priors with proposals is the possibility to train the network with batches containing 2D images sampled from different 3D volumes, which is much more challenging in direct optimization approaches like the penalty method. It also enables incorporating more complex constraints and regularization priors in the learning process, a promising research direction that we will investigate in future work.

\section*{Acknowledgements}  

We acknowledge the support of the Natural Sciences and Engineering Research Council of Canada (NSERC), and thank NVIDIA corporation for supporting this work through their GPU grant program.

\section*{References}

\bibliography{mybibfile}

\end{document}